\documentclass[11pt]{article}

\usepackage[]{ACL2023}

\usepackage{times}
\usepackage{latexsym}
\usepackage{xspace}

\usepackage[T1]{fontenc}

\usepackage[utf8]{inputenc}

\usepackage{microtype}

\usepackage{inconsolata}
\usepackage{comment}
\usepackage{amsmath}
\usepackage{amssymb}
\usepackage{multirow}
\usepackage{graphicx}
\usepackage{colortbl}
\usepackage{bbding}
\newcommand{\method}{FinTrust\xspace}
\title{Measuring Consistency in Text-based Financial Forecasting Models}

\author{Linyi Yang$^{1,2*}$, Yingpeng Ma$^{1,2*}$,  Yue Zhang$^{1,2\dagger}$\\
  $^{1}$Institute of Advanced Technology, Westlake Institute for Advanced Study \\
  $^{2}$School of Engineering, Westlake University \\
  \texttt{yanglinyi,mayingpeng,yuezhang@westlake.edu.cn} \\ \\}
\begin{document}
\maketitle

\def\thefootnote{*}\footnotetext{ Equal contribution. Yingpeng Ma did this work during his internship at Westlake University.}
\begin{abstract}

Financial forecasting has been an important and active area of machine learning research, as even the most modest advantage in predictive accuracy can be parlayed into significant financial gains. Recent advances in natural language processing (NLP) bring the opportunity to leverage textual data, such as earnings reports of publicly traded companies, to predict the return rate for an asset. However, when dealing with such a sensitive task, the consistency of models -- their invariance under meaning-preserving alternations in input -- is a crucial property for building user trust. Despite this, current financial forecasting methods do not consider consistency. To address this problem, we propose \method, an evaluation tool that assesses logical consistency in financial text. Using \method, we show that the consistency of state-of-the-art NLP models for financial forecasting is poor. Our analysis of the performance degradation caused by meaning-preserving alternations suggests that current text-based methods are not suitable for robustly predicting market information. All resources are available at \url{https://github.com/yingpengma/FinTrust}.

\end{abstract}

\section{Introduction}

NLP techniques have been used in various financial forecasting tasks, including stock return prediction, volatility forecasting, portfolio management, and more \cite{Ding14,Ding15,Qin19,xing2020financial,du2020stock,yang2020html,sawhey2020mm}. Despite the increased performance of NLP models on financial applications, there has been pushback questioning their trustworthiness, and robustness \cite{chen2022fintech,li2022simpler}. Recently, the causal explanation has been viewed as one of the promising directions for measuring the robustness and thus improving the transparency of models \cite{stolfo2022causal,feder2022causal}. Among them, consistency has been viewed as a crucial feature, reflecting the systematic ability to generalize in semantically equivalent contexts and receiving increasing attention in tasks such as text classification and entailment \cite{jin2020bert,jang2022becel}.

\begin{figure}[t]
\centering 
{%
\includegraphics[width=.42\textwidth]{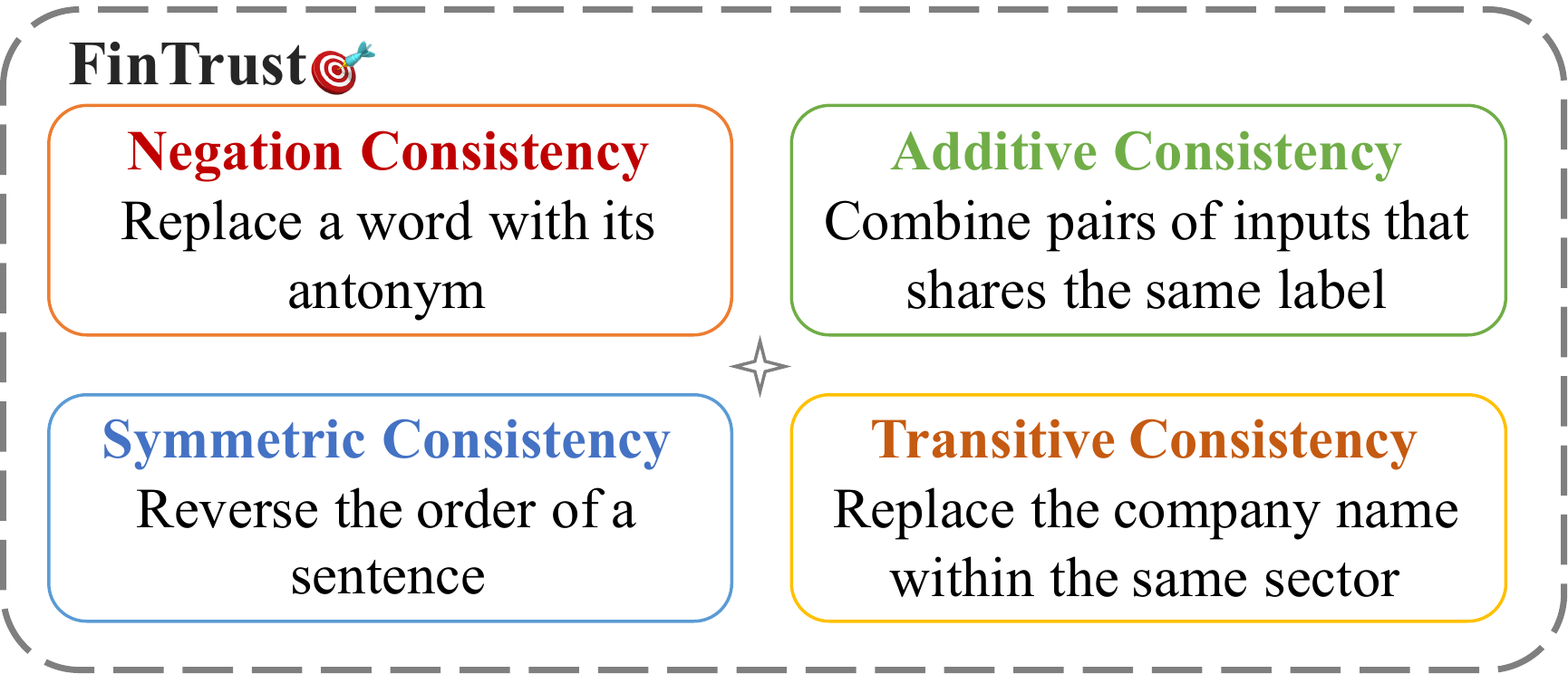} 
}
\caption{Examples of four consistency transformations used in \method.} 
\label{fig:intro}
\end{figure}

Previous text-based financial forecasting methods have mostly considered stock movement prediction based on various sources of data, including financial news \cite{Xu18,Zhang18News}, analyst reports \cite{Kogan2009report,Rekabsaz2017report}, and earnings conference calls \cite{Qin19,Keith19,li2020maec,chen2021opinion}. While most work evaluates their methods using accuracy and profit gains based on the final outcome in the market \cite{sawhney-etal-2021-multimodal,yang2022numhtml}, consistency evaluation remains largely unexplored. The only exception \cite{chuang-yang-2022-buy} focuses on evaluating the implicit preferences in Pre-trained Language Models (PLMs) but not the consistency in predictive models. The lack of evaluation in behavior consistency, an important characteristic of human decisions, hinders the deployment of financial forecasting models in real-world scenarios. 

The main objective of this work is to explore a wholistic measure for stock movement prediction, integrating consistency as a criterion of trustworthiness. To this end, we define \emph{behavior consistency} of text-based models in the financial domain. Regarding the intrinsic characteristics of financial text data, we consider four types of logical consistency tests. As shown in Figure \ref{fig:intro}, these transformations include Negation Consistency, Symmetric Consistency, Additive Consistency, and Transitive Consistency. Taking negation consistency as an example, given an input \textit{"the cost of raw materials has been greatly decreased"}, if the token \textit{"decreased"} is changed to \textit{"increased"}, the model prediction is expected to be flipped accordingly. 

Based on the above logical transformations, we introduce \method , a new evaluation tool that enables researchers to measure consistency in PLMs and text-based financial forecasting models. Using \method , we design three tasks to investigate the influence of these logical transformations. First, we assess implicit preference in PLMs such as BERT \cite{devlin2018bert} and FinBERT \cite{yang2020finbert}, especially for economic words. Second, we measure the accuracy of stock movement prediction on a real-world earnings call dataset after the meaning-preserving modifications. Finally, we propose a realistic trading simulation to see if simple meaning-preserving modifications can wipe out positive returns. 

Experiments on several baseline models, including previous best-performing architectures \cite{Ding15,Qin19,yang2020html} and the machine learning classifier \cite{chen2015xgboost} show that all current methods exhibit a significant decline in the performance of stock movement predictions when evaluating on \method compared to their original results. Notably, some models demonstrate a level of accuracy that is even lower than that of a random guess after undergoing logical consistency transformation, and most methods fail to surpass the performance of the simplest Buy-all strategy in the trading simulation. These results suggest that existing text-based financial models have robustness and trustworthiness issues, which can limit their use in practical settings.

To our knowledge, \method  is the first evaluation tool for probing if the relatively accurate stock movement prediction is based on the right logical behavior. We release our tool and dataset at Github\footnote{ \url{https://github.com/yingpengma/FinTrust}}, which can assist future research in developing trustworthy FinNLP methods.

\section{Related Work}

\textbf{Text-based Financial Forecasting.} A line of work has leveraged event-based neural networks based on financial news for predicting the stock movement of S\&P 500 companies \cite{Ding14,Ding15,Xu18}. By taking advantage of recent advances in NLP, recent work has shown potential in predicting stock price movements using PLMs, BERT \cite{devlin2018bert}, and FinBERT \cite{araci2019finbert,yang2020finbert}, with rich textual information from social media and earnings conference calls \cite{liu2013estimation,xing2020financial,chen2021evaluating}. The considerable PLMs mainly include BERT and FinBERT. While BERT is trained on corpora from fairly general domains, FinBERT is trained on financial corpora, including earnings conference calls and analyst reports, under the same architecture as BERT. Although implicit stock market preference is in the masked token predictions task, the implicit preference has been under-explored using a logical behavior test.

In addition to building pre-trained models specially trained for financial domains, researchers have recently proposed myriad neural network architectures aimed at more accurate predictions to produce profitable gains including financial risk (volatility) and return predictions. For example, researchers \cite{Qin19,yang2020html,sawhney2020multimodal} have considered predicting the volatility of publicly traded companies based on multi-model earnings conference call datasets. Also, \citet{Xu18, duan2018learning, yang2018explainable,feng19} leverage different textual data sources for predicting the stock movement based on the daily closing price. Unfortunately, despite the alarm over the reliance of machine learning systems on spurious patterns that have been found in many classical NLP tasks, the topic of text-based financial forecasting lacks a systematical evaluation regarding the robustness analysis from either an adversarial or consistency perspective. To this end, we present the first critical investigation of popular benchmarks by using \method from the consistency perspective.

\begin{figure*}[t]
    \centering
    \includegraphics[width=.98\linewidth]{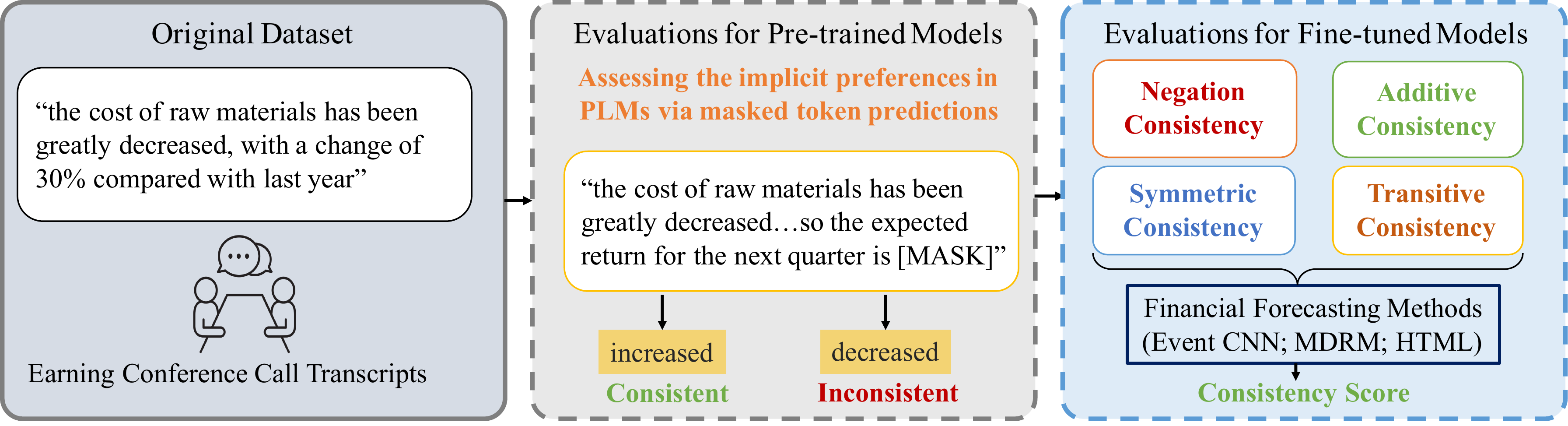}
    \caption{Pipelines of \method consist of evaluating pre-trained features (e.g., BERT and FinBERT) and fine-tuned text-based financial forecasting models.}
    \label{fig:2}
\end{figure*} 

\noindent\textbf{Consistency Measurement.} The inductive bias of machine learning systems is greatly affected by the patterns in training data due to the nature of inductive reasoning. While a flurry of research has highlighted this issue \cite{gururangan2018annotation,srivastava2020robustness,garg2020bae,ICLR20Counterfact}, recent work \citet{jang2022becel} shows that possible artefacts in data are more influential than the model design when leading to the problem of lacking trustworthiness. Thus, assessing the influence of data artefacts, such as consistency, becomes a crucial problem for trustworthy NLP. \citet{elazar2021measuring} study the consistency of PLMs (e.g., BERT, ALBERT, and RoBERTa) with regard to their knowledge extraction ability and conclude that the consistency of these models is generally low. \citet{chuang-yang-2022-buy} aim to raise awareness of potential implicit stock preferences based on the finding that consistent implicit preference of the stock market exists in PLMs at the whole market.

In addition to evaluating preferences in PLMs, previous methods also attempt to evaluate the consistency of models in downstream NLP tasks, such as visual question answering \cite{ribeiro2018semantically}, QA \cite{jia2017adversarial,ribeiro2019red,gan2019improving,asai2020logic}, named entity recognition \cite{jia-etal-2019-cross,wang-henao-2021-unsupervised}, and natural language inference \cite{naik-etal-2018-stress,hossain-etal-2020-analysis,camburu2020make,sinha-etal-2021-unnatural}. Besides, \citet{ribeiro-etal-2020-beyond} consider using consistency for building the behavioural testing benchmark beyond accuracy. Surprisingly, these discussions have not yet been extended to text-based financial forecasting models, which require strong robustness to assist decision-making in the financial market, with the exception of our work.

\section{Method}
\label{sec:bibtex}

We define the pipeline of \method in Figure \ref{fig:2}. For text-based financial models, there are two salient components, namely text representations and financial behavior. For the former, using PLMs has become a dominant approach, improving the quality of text representations in many domains. For the latter, various neural models can be built on PLMs. Correspondingly, we have two setups in the consistency evaluation, representation (Setup 1) and behavior (Setup 2), respectively.

\noindent\textbf{Setup 1.} In the first stage, we assess the implicit preferences in PLMs via masked token predictions. In particular, we first mask a predictable word from the original input extracted from earning conference call transcripts, such as \textit{"the cost of raw materials has been greatly decreased...so the expected return for the next quarter is [MASK]"}. Then, we predict the masked token using PLMs and compare the probability of predicting "increased" and "decreased" for contexts from different transcripts. A higher probability of predicting "increased" would indicate that the given PLM hold logical consistency with human predictions. Conversely, it suggests that the prediction of the PLM may be influenced by spurious patterns such as favoritism towards a particular stock.

\noindent\textbf{Setup 2.} We evaluate text-based financial forecasting models after fine-tuning PLMs on a popularly used earnings conference call dataset \cite{Qin19}. However, the consistency measurement faces significant challenges in defining the relationship between two texts, particularly when the text is a long transcript with complex logical connections, such as earnings conference call transcripts. Incorrectly defining this relationship can render consistency judgments meaningless. In line with prior research \cite{jang2022becel}, we develop four logical consistency transformations customized for financial text in this work.  By meaning-preserving altering the original text, we ensure generated samples have a logical relationship to the original text, thus ensuring the consistency judgment is meaningful. Below we define our text level consistency transformation first (Sec 3.1), before introducing the financial tasks for behavior study (Sec 3.2) and a wholistic metric (Sec 3.3) to integrate performance and trustworthiness.


\subsection{Logical Consistency Transformations on Text Data}

In \method, four logical consistency transformation approaches are defined to evaluate if the model maintains the same logical behavior as humans, representing the consistency in text-based financial forecasting models. 

\textbf{Negation consistency} refers to the ability of a model to generate converse predictions for texts with opposite meanings, i.e. $f(x) = positive \Leftrightarrow f(\neg x) = negative$, where $x$ is the input transcript, $f(x)$ represents the output of the model, a "positive" outcome means the stock price will increase, and a "negative" outcome means the stock price will decrease. $\neg x$ is a negation consistency transformed test example flipped through predetermined rules based on the bi-grams of the most frequent words and their antonyms. We achieve this by splitting the dataset at the sentence level and flipping the meanings of sentences. Given an input \textit{"the cost of raw materials has been greatly decreased, with a change of 30\% compared with last year"}, its counterpart can be \textit{"the cost of raw materials has been greatly increased, with a change of 30\% compared with last year"}. In the financial market, a significant cost reduction may lead to optimism about the company's future prospects and an increase in stock price. Only when the model can give the correct predictions for both pairs of testing data we consider that the model is consistent with non-contradictory predictions. Otherwise, it is considered to lack negation consistency.

\textbf{Symmetric consistency} is the property of a model where the order of the inputs does not affect the output. It is defined as $f(S_{p1}, S_{p2})=f(S_{p2}, S_{p1})$, where $S$ is a sentence in the transcript, $S_{pi}$ represents the part $i$ of the sentence. This can be tested by reordering the segments of each sentence in the transcript and comparing the predictions before and after the reordering. For example, given the sentence \textit{"the cost of raw materials has been greatly decreased, with a change of 30\% compared with last year"}, if the prediction is reversed after reordering it to \textit{"with a change of 30\% compared with last year, the cost of raw materials has been greatly decreased"}, then the model is regarded as lacking symmetric consistency.

\textbf{Additive consistency} refers to the property of a model 
 to predict the stock movement based on the combination of two inputs, $x$ and $y$ that share the same label. The model is expected to hold the same prediction for $x$, $y$, and the concatenation of those inputs $x+y$. If the model produces different predictions for the above three kinds of inputs, it can be regarded as lacking additive consistency. For example, if a model gives a positive prediction for the sentence \textit{"the cost of raw materials has been greatly decreased, with a change of 30\% compared with last year"}, and also gives a positive prediction for the sentence \textit{"we believe that our products can bring convenience to everyone's life"}, then it should also make a positive prediction for the combined sentences after the concatenation.

\textbf{Transitive consistency} refers to the ability of a model where the perceived sentiment of a company should be reflected in the performance of the top-valued company in the same industry. It can be expressed as $f(x)=f(x')$, where $x'$ represents transitive consistency transformed text. Specifically, for transcripts of a particular company, the top-valued company in the same industry is identified and its name is denoted as "company\_name". Then occurrences of words such as "we" and "our" are replaced with "company\_name" and "company\_name's" respectively. For example, if the corresponding sector of the company is "Information Technology" and the top-valued company in the S\&P 500 is Apple Inc., a sentence such as "\textit{we believe that our products can bring convenience to everyone's life}" will be transformed to "\textit{Apple Inc. believe that Apple Inc.'s products can bring convenience to everyone's life}" after transitive consistency transformation. Again, we calculate the consistency of models by considering the non-contradictory predictions over transitive instances.

\subsection{Prediction Tasks in FinTrust}

\textbf{Consistency Measurement in PLMs.} To better assess the implicit preference in PLMs, we extend the previous cloze-style prompts used in assessing stock market preference \cite{chuang-yang-2022-buy} by considering logical changes rather than simply predicting the masked token in the input. This is crucial as if PLMs are biased, the fine-tuned model's predictions based on features learned by PLMs could be further influenced by spurious preference tendencies, which would negatively impact the effect of financial forecasting. 

\noindent\textbf{Stock Prediction Task.} Following previous studies \cite{Ding15,duan2018learning,sawhey2020mm}, we treat the stock movement prediction as a binary classification problem, where the model predicts whether the daily closing price of a given asset will increase or decrease over the next \emph{n} days (\emph{n}=3, 7, 15, 30) based on the content of earnings call transcripts. The output is either ``increase'' (positive) or ``decrease'' (negative).

\noindent\textbf{Trading Simulation Task.} We use the predictions to determine whether to buy or sell a stock after n days. For example, if the model predicts that the stock price would increase from day $d$ to day $d+30$, we would buy the stock on day $d$ and sell it on day $d+30$. Otherwise, we execute a short sell. The previous work \citet{sawhney2020multimodal} simulates the trade of one hand for each stock, which allows for the potential offset of multiple forecast failures if one stock is more valuable. However, this approach is unfair under specific situations since each prediction and trade are treated equally and thus will lose the balance between trades. Therefore, we invest the same amount of money in each stock and calculate the profit ratio instead of the cumulative profit. This method does not affect the calculation of the Sharpe Ratio and allows us to explore the impact of financial forecasting consistency on performance and profitability. Notably, we do not consider the transaction cost in accordance with previous work \cite{sawhney2020multimodal}. 

\subsection{Wholistic Evaluation Metrics}
We introduce the predictive evaluation metrics and the novel consistency evaluation metrics as elaborated below.

\noindent\textbf{Predictive Evaluations.} For stock prediction, we use three metrics to measure performance: Accuracy, F1 score, and Matthews correlation coefficient (MCC). These metrics are calculated as follows:

\begin{equation}
\small
{F1 = \frac{2 \times precision \times recall}{precision + recall}}
\end{equation} For a given confusion matrix:

\begin{equation}
\small
{Accuracy = \frac{tp + tn} {tp + tn + fp + fn}}
\end{equation}

\begin{equation}
\small
{MCC = \frac{tp \times tn - fp \times fn}{\sqrt{(tp + fp)(tp + fn)(tn + fp)(tn + fn)}}}
\end{equation} We use both Profit Ratio and Sharpe Ratio for the trading simulation task as performance indicators. Return $R$, and investment $I$ is involved in calculating the Profit Ratio.

\begin{equation}
\small
{Profit Ratio = \frac{R}{I} }
\end{equation} The Sharpe Ratio measures the performance of an investment by considering the average return $R_x$, risk-free return $R_f$, and standard deviation of the investment $\sigma(R_x)$.

\begin{equation}
\small
{Sharpe Ratio = \frac{R_x-R_f}{\sigma(R_x)} }
\end{equation}

\noindent\textbf{Consistence Evaluations.} Based on logical transformations, we propose the consistency evaluation metrics of consistency, aiming to measure text-based financial forecasting models from a consistency perspective as a complementary metric to accuracy. Assuming that $C$ is a set of four logical consistencies. To begin with, we define the consistency score ($Consis$), elaborated as follows:
\begin{equation}
{Consis = \frac{\sum_{i=1}^{\left| C \right|}C_i}{\left| C \right|}} 
\end{equation} where the $C$ set contains Negation consistency $Consis^N$, Symmetric consistency $Consis^S$, Additive consistency $Consis^A$, Transitive consistency $Consis^T$. We give the formal definition of those four metrics, respectively. The consistency of $Consis^N$ is calculated as:

\begin{equation}
{Consis^N = \frac{\sum_{i=1}^{\left| D \right|} 
\left\{
\begin{array}{ll}
0\enspace (f(x_i) = f(x_i^N))
\\
1\enspace (f(x_i) \neq f(x_i^N))
\end{array}
\right. 
}{\left| D \right|}} 
\end{equation} where $D$ is the original test set, $x_i$ is the test sample in the original test set, i.e. $x_i \in D$. $x_i^N$ is the new test sample obtained by negation consistency transformation on $x_i$, and $f(x)$ is the prediction of the model (positive or negative) for the input $x$. In terms of the symmetric, additive, and transitive transformations, the value equals 0 when $f(x_i) \neq f(x_i^N)$ while equals 1 when $f(x_i) = f(x_i^N)$.

\section{Experiments}
\label{sec:bibtex}
We first evaluate the explicit preferences in PLMs. Then we assess the ability of text-based models to make consistent predictions on the stock movement and finally test the profitability of these predictions using a trading simulation.

\subsection{Dataset}

\textbf{Earnings Call Data.} We use the publicly available Earning Conference Calls dataset by \cite{Qin19}, which includes transcripts of 576 earnings calls from S\&P 500 companies listed on the American Stock Exchange, obtained from the Seeking Alpha website. It also includes the meta-information on the company affiliations and publication dates.

\noindent\textbf{Financial Market information.} We also collect historical price data (closing price) for the traded companies listed in S\&P 500 from Yahoo Finance for the period from January 1, 2017, to January 31, 2018. This data was used to calculate the label of stock price movement and profitability.

\noindent\textbf{Data Processing.} Following \cite{Qin19,yang2020html}, we split the dataset into mutually exclusive train/validation/test sets in a 7:1:2 ratio in chronological order to ensure that future information is not used to predict past price movements. We also construct logical consistency datasets based on the original test set using the above-mentioned four logical consistency transformations. The size of our evaluation dataset is four times the size of the original one since we ensure that each sample in the original test set corresponds to four logical consistency test samples. To facilitate future research, we release our dataset and the evaluation toolkit in \textbf{FinTrust}.

\subsection{Models}

\textbf{Representation Models.} We conduct experiments on popular PLMs, including BERT \cite{devlin2018bert}, RoBERTa \cite{liu2019roberta}, DistilBERT \cite{sanh2019distilbert}, and FinBERT \cite{yang2020finbert}. The vocabulary of FinBERT is different from the others as it contains domain-specific terms in the financial market, including company names.

\noindent\textbf{Predictive Models.} Regarding the forecasting models, we evaluate several baselines, including the traditional machine learning and state-of-the-art transformer-based methods, detailed as follows.  

\begin{itemize}
\item \textbf{HTML:} \citet{yang2020html} propose a hierarchical transformer-based framework to address the problem of processing long texts in earnings call data. It utilizes a pre-trained WWM-BERT-Large model to generate sentence representations as inputs for the model.
\item \textbf{MRDM:} \citet{Qin19} propose the first method to treat volatility prediction as a multi-modal deep regression problem, building benchmark results and introducing the earnings conference call dataset.
\item \textbf{Event:} \citet{Ding15} adapt Open IE for event-based stock price movement prediction, extracting structured events from large-scale public news without manual efforts. 
\item \textbf{XGBoost:} \citet{chen2015xgboost} propose a gradient-boosting decision tree known as the classical machine learning baseline.
\end{itemize}

\begin{table}[]
\centering
\small
\begin{tabular}{cc|ccc}
\hline
\textbf{PLM} &\textbf{Params} & \textbf{Neg} & \textbf{Pos} & \textbf{Consistency} \\ \hline
BERT-base & 110M   & +             & +             & 71.33\%     \\
BERT-base & 110M   & +             & -             & 55.87\%     \\
BERT-base & 110M   & -              & +             & 86.79\%     \\ \hline
BERT-large & 340M     & +             & +             & 75.67\%     \\
BERT-large & 340M     & +             &  -             & 67.60\%     \\
BERT-large & 340M     &  -             & +             & 83.74\%     \\ \hline
RoBERTa-base & 125M        & +             & +             & 77.79\%     \\
RoBERTa-base & 125M        & +             &  -             & 69.17\%     \\
RoBERTa-base & 125M        &   -            & +             & 86.40\%     \\ \hline
RoBERTa-large & 355M  & +             & +             & \textbf{82.70\% }    \\
RoBERTa-large & 355M  & +             &   -            & \textbf{76.67\%}     \\
RoBERTa-large & 355M  &   -            & +             & \textbf{88.72\%}    \\ \hline
FinBERT & 110M  & +             & +             & 72.40\%     \\
FinBERT & 110M  & +             &   -            & 56.27\%     \\
FinBERT & 110M  &   -            & +             & 88.53\%     \\ \hline
DistilBERT & 66M      & +             & +             & 70.13\%     \\
DistilBERT & 66M      & +             &   -            & 57.92\%     \\
DistilBERT & 66M      &   -            & +             & 82.33\%     \\
\hline
\end{tabular}%
\caption{The results of the consistency measurement in PLMs via masked token predictions, splitting by negative and positive token predictions. `+' denotes that the attitude of the word with the specific polarity will be predicted while `-' means that we do not consider tokens with a specific polarity.}
\label{tab:evalplm}
\end{table}

\section{Results and Discussion}
We report the results of three tasks defined in Section 3.2 and the consistency score calculated by the consistency evaluation metrics in this section. Furthermore, we present extensive ablation studies and discussions to support in-depth analyses of each component in FinTrust.
\begin{table*}[t]
\centering
\resizebox{\textwidth}{!}{%
\begin{tabular}{c|l|llll|l|llll|l|llll}
\hline
\textbf{Metrics} &
  \multicolumn{5}{c}{\textbf{ACC}} &
  \multicolumn{5}{c}{\textbf{F1}} &
  \multicolumn{5}{c}{\textbf{MCC}} \\  \hline
Period &
  \multicolumn{1}{c}{Avg} &
  \multicolumn{1}{c}{3} &
  \multicolumn{1}{c}{7} &
  \multicolumn{1}{c}{15} &
  \multicolumn{1}{c}{30} &
  \multicolumn{1}{c}{Avg} &
  \multicolumn{1}{c}{3} &
  \multicolumn{1}{c}{7} &
  \multicolumn{1}{c}{15} &
  \multicolumn{1}{c}{30} &
  \multicolumn{1}{c}{Avg} &
  \multicolumn{1}{c}{3} &
  \multicolumn{1}{c}{7} &
  \multicolumn{1}{c}{15} &
  \multicolumn{1}{c}{30} \\ \hline
HTML &
  \textbf{0.546} &
  0.442 &
  0.531 &
  0.566 &
  0.646 &
  \textbf{0.671} &
  0.571 &
  0.619 &
  0.713 &
  0.780 &
  \textbf{0.078} &
  0.052 &
  0.056 &
  0.032 &
  0.175 \\
+\method &
  \textbf{0.521}$\downarrow$ &
  0.465$\uparrow$ &
  0.527$\downarrow$ &
  0.529$\downarrow$ &
  0.564$\downarrow$ &
  \textbf{0.647}$\downarrow$ &
  0.608$\uparrow$ &
  0.629$\uparrow$ &
  0.648$\downarrow$ &
  0.703$\downarrow$ &
  \textbf{0.040}$\downarrow$ &
  0.019$\downarrow$ &
  0.058$\uparrow$ &
  0.019$\downarrow$ &
  0.063$\downarrow$ \\ \hline
MRDM &
  \textbf{0.555} &
  0.504 &
  0.513 &
  0.584 &
  0.619 &
  \textbf{0.670} &
  0.541 &
  0.663 &
  0.722 &
  0.754 &
  \textbf{0.059} &
  0.079 &
  0.007 &
  0.107 &
  0.044 \\
+\method&
  \textbf{0.504}$\downarrow$ &
  0.465$\downarrow$ &
  0.511$\downarrow$ &
  0.507$\downarrow$ &
  0.535$\downarrow$ &
  \textbf{0.622}$\downarrow$ &
  0.569$\uparrow$ &
  0.667$\uparrow$ &
  0.578$\downarrow$ &
  0.674$\downarrow$ &
  \textbf{0.017}$\downarrow$ &
  -0.024$\downarrow$ &
  0.038$\uparrow$ &
  0.032$\downarrow$ &
  0.023$\downarrow$ \\ \hline
Event &
  \textbf{0.542} &
  0.416 &
  0.522 &
  0.593 &
  0.637 &
  \textbf{0.694} &
  0.582 &
  0.682 &
  0.736 &
  0.776 &
  \textbf{0.122} &
  0.078 &
  0.097 &
  0.189 &
  0.123 \\
+\method&
  \textbf{0.512}$\downarrow$ &
  0.447$\uparrow$ &
  0.504$\downarrow$ &
  0.529$\downarrow$ &
  0.569$\downarrow$ &
  \textbf{0.656}$\downarrow$ &
  0.598$\uparrow$ &
  0.663$\downarrow$ &
  0.658$\downarrow$ &
  0.705$\downarrow$ &
  \textbf{0.006}$\downarrow$ &
  -0.032$\downarrow$ &
  -0.023$\downarrow$ &
  0.013$\downarrow$ &
  0.068$\downarrow$ \\ \hline
XGB &
  \textbf{0.515} &
  0.434 &
  0.487 &
  0.584 &
  0.558 &
  \textbf{0.561} &
  0.448 &
  0.500 &
  0.641 &
  0.653 &
  \textbf{0.018} &
  -0.093 &
  -0.027 &
  0.147 &
  0.043 \\
+\method&
  \textbf{0.507}$\downarrow$ &
  0.462$\uparrow$ &
  0.502$\uparrow$ &
  0.531$\downarrow$ &
  0.531$\downarrow$ &
  \textbf{0.545}$\downarrow$ &
  0.456$\uparrow$ &
  0.518$\uparrow$ &
  0.584$\downarrow$ &
  0.622$\downarrow$ &
  \textbf{-0.004}$\downarrow$ &
  -0.076$\uparrow$ &
  -0.002$\uparrow$ &
  0.045$\downarrow$ &
  0.014$\downarrow$ \\ \hline
\end{tabular}%
 }
\caption{Performance and robustness evaluation of stock movement prediction for multiple baselines using FinTrust. Significant performance decay has been observed on all methods using the Student T-test over 10 times run, p$<$0.05.}
\label{tab:table2}
\end{table*}

\subsection{Predictive Results}
\noindent\textbf{Consistency Measurement in PLMs.} The results of explicit preferences in PLMs are presented in Table \ref{tab:evalplm}. In general, we find that all PLMs exhibited relatively low consistency, ranging from 70.13\% to 82.7\%, which falls significantly short of the level of robustness expected in the financial market.  Also, we observe that PLMs typically demonstrated lower consistency when tested on negative tokens than positive tokens (on average 63.91\% -- negative vs. 86.09\% -- positive). This suggests that popular PLMs tend to exhibit stereotypes when predicting negative tokens. 

From a model-level perspective, our results indicate that FinBERT, which utilizes a domain-specific training corpus during the pre-training phase, can slightly improve consistency compared to BERT-base. Besides, we show that the increase in parameter size brings significant benefits for improving consistency, given that BERT-large and RoBERTa-large both outperform their base-sized versions (75.67\% vs. 71.33\% -- BERT; 82.70\% vs.77.79\% --RoBERTa). In particular, RoBERTa-achieves the highest consistency across three settings, indicating its high robustness. In contrast, DistilBERT achieves the lowest consistency.


\noindent\textbf{Stock Movement Prediction.} The results of stock movement prediction over text-based financial forecasting models are shown in Table \ref{tab:table2}. We evaluate multiple baselines by comparing the results of models on the original test set to the results tested on transformed datasets (shown as +FinTrust). It is noteworthy that the accuracy of some predictions is even lower than that of random guess, especially for the short-time prediction (n=3). Furthermore, we demonstrate that the effect of logical consistency transformations on traditional performance indicators varies depending on the time period, but the average performance of all models decreased significantly over three metrics. In particular, models show extraordinary vulnerability when it comes to predicting the long-term stock return (n=15 and 30), as transformations in all settings decrease accuracy when the time period is 15 and 30 days. 

From the model perspective, regarding the ratio of performance decay, XGBoost is the least impacted, and MRDM is the most affected. This can be because traditional machine learning models, such as XGBoost, have fewer parameters than deep learning models and are therefore less affected by artefacts. Despite this, the accuracy on FinTrust achieved by models is only slightly more accurate than the random guess (e.g., \textbf{0.504} on MRDM, \textbf{0.507} on XGBoost). The vulnerability of these models, including state-of-the-art methods, hinders the deployment of NLP systems in the real financial market and should be taken more seriously.

\begin{table}[t]
\centering
\small
\begin{tabular}{c l l}
\hline
\textbf{Strategy} & \textbf{Profit Ratio} & \textbf{Sharpe Ratio} \\ \hline
HTML           & \hspace{1.3em}3.752        &\hspace{1.3em}0.266        \\
+ FinTrust      & \hspace{1.3em}3.359$\downarrow$       &\hspace{1.3em}0.229$\downarrow$       \\
$\Delta$$\downarrow$       & \hspace{1.3em}-10\%        &\hspace{1.3em}-14\%        \\ \hline
Event          &\hspace{1.3em}3.720        &\hspace{1.3em}0.263        \\
+ FinTrust       & \hspace{1.3em}3.535$\downarrow$       &\hspace{1.3em}0.245$\downarrow$       \\
$\Delta$$\downarrow$            & \hspace{1.3em}-5\%        &\hspace{1.3em}-7\%        \\ \hline
MRDM           & \hspace{1.3em}3.495        &\hspace{1.3em}0.241        \\
+ FinTrust      & \hspace{1.3em}2.384$\downarrow$       &\hspace{1.3em}0.138$\downarrow$       \\
$\Delta$$\downarrow$            & \hspace{1.3em}-32\%        &\hspace{1.3em}-43\%        \\ \hline
XGB            & \hspace{1.3em}-0.515       &\hspace{1.3em}-0.126       \\
+ FinTrust       & \hspace{1.3em}0.296$\uparrow$       &\hspace{1.3em}0.032$\uparrow$       \\
$\Delta$$\uparrow$        & \hspace{1.3em}158\%        & \hspace{1.3em}75\%        \\ \hline
Buy-all        & \hspace{1.3em}3.681        & \hspace{1.3em}0.259        \\
Random         & \hspace{1.3em}-0.271       &\hspace{1.3em}-0.105       \\
Short-sell-all & \hspace{1.3em}-3.681       &\hspace{1.3em}-0.259     \\  \hline
\end{tabular}%
\caption{Performance on the trading simulation. `+FinTrust' represents the performance using the input after the transformation.}
\label{tab:my-table3}
\end{table}

\noindent\textbf{Trading Simulation.} We compare three simple trading strategies (Buy-all, Short-sell-all, and Random) with four baselines. The results are shown in Table \ref{tab:my-table3}. It can be seen that HTML and Event have higher yields and can exceed simple trading strategies. However, after conducting consistency transformations, positive returns of these two methods are much reduced, even lower than the simple Buy-all strategy. Methods such as MRDM and XGBoost gain lower returns than Buy-all, with MRDM experiencing the highest drop of about \textbf{32\%-43\%}. Even though the returns of XGBoost improved significantly after the transformations, it still remained much lower than the Buy-all strategy and the other three baselines. Hence, we contend that the increase in XGBoost's returns does not have a strong reference value. We conclude that most methods show unacceptably poor performance caused by lacking consistent logical behavior.



\begin{table}[t]
\centering
\small
\begin{tabular}{cc|cccc}
\hline
\multicolumn{2}{c|}{\textbf{Period}}   & \textbf{3}     & \textbf{7}     & \textbf{15}    & \textbf{30}    \\ \hline
              & AVG         & \textbf{0.730} & \textbf{0.739} & 0.644 & \textbf{0.692} \\
              & Add         & 0.903 & 0.947 & 0.664 & 0.805 \\
Event         & Neg         & 0.106 & 0.035 & 0.044 & 0.018 \\
              & Sym         & 0.947 & 0.982 & 0.929 & 0.973 \\
              & Tra         & 0.965 & 0.991 & 0.938 & 0.973 \\ \hline
              & AVG         & 0.699 & 0.628 & \textbf{0.688}& 0.684 \\
              & Add         & 0.894 & 0.655 & 0.876 & 0.743 \\
HTML          & Neg         & 0.115 & 0.212 & 0.177 & 0.009 \\
              & Sym         & 0.894 & 0.796 & 0.841 & 0.991 \\
              & Tra         & 0.894 & 0.850 & 0.858 & 0.991 \\ \hline
              & AVG         & 0.597 & 0.706 & 0.524 & 0.650 \\
              & Add         & 0.664 & 0.894 & 0.301 & 0.735 \\
MRDM          & Neg         & 0.248 & 0.062 & 0.053 & 0.053 \\
              & Sym         & 0.655 & 0.894 & 0.805 & 0.885 \\
              & Tra         & 0.823 & 0.973 & 0.938 & 0.929 \\ \hline
              & AVG         & 0.566 & 0.595 & 0.593 & 0.653 \\
              & Add         & 0.522 & 0.487 & 0.496 & 0.504 \\
XGB           & Neg         & 0.071 & 0.133 & 0.124 & 0.354 \\
              & Sym         & 1.000 & 0.973 & 0.973 & 0.991 \\
              & Tra         & 0.673 & 0.788 & 0.779 & 0.761 \\ \hline
\end{tabular}%
\caption{The consistency score calculated by $Consis$.}
\label{tab:consis}
\end{table}

\subsection{Consistency Score} 

\noindent\textbf{Results.} We show the results of the consistency score (defined in Section 3.4) in Table \ref{tab:consis}. It can be seen that Event has the highest consistency score (\emph{Consis}) and XGBoost has the lowest \emph{Consis}. Regarding the average consistency over four transformations, Event achieves three of the four highest consistency scores. XGBoost tends to make contradictory predictions in terms of the lowest scores in three settings. Additionally, all methods perform poorly on negation consistency, consistent with findings in the PLMs evaluation (Table \ref{tab:evalplm}). 

\noindent\textbf{Correlation Analysis.} We examine the correlation between the indicators of consistency and accuracy. Importantly, we find that our consistency score does not align with traditional performance indicators such as accuracy, evidenced by the fact that the most consistent model (Event) is not necessarily the highest in accuracy (HTML). The overall Pearson correlation coefficient between the consistency score and accuracy is only 0.314, indicating a low-level correlation. This suggests that the proposed consistency score can be used as a complementary evaluation metric for accuracy in future research on text-based financial forecasting.

\subsection{Discussion}

\noindent\textbf{Human Evaluation.} To assess the effectiveness of our consistency transformation method in preserving the original meaning, we conduct a human annotation study. Two annotators are  employed from the author list and be required to label each sample and its four consistency transformations. Both of them received an advanced degree in computer science. The Inter-Annotator Agreement score is calculated to be 0.98, based on an evaluation of 40 samples and their 160 transformed samples. The average consistency score for human annotators is 0.975, indicating that our method successfully preserves the original meaning in most cases. 


\begin{figure}[t]
    \centering
    \includegraphics[width=.98\linewidth]{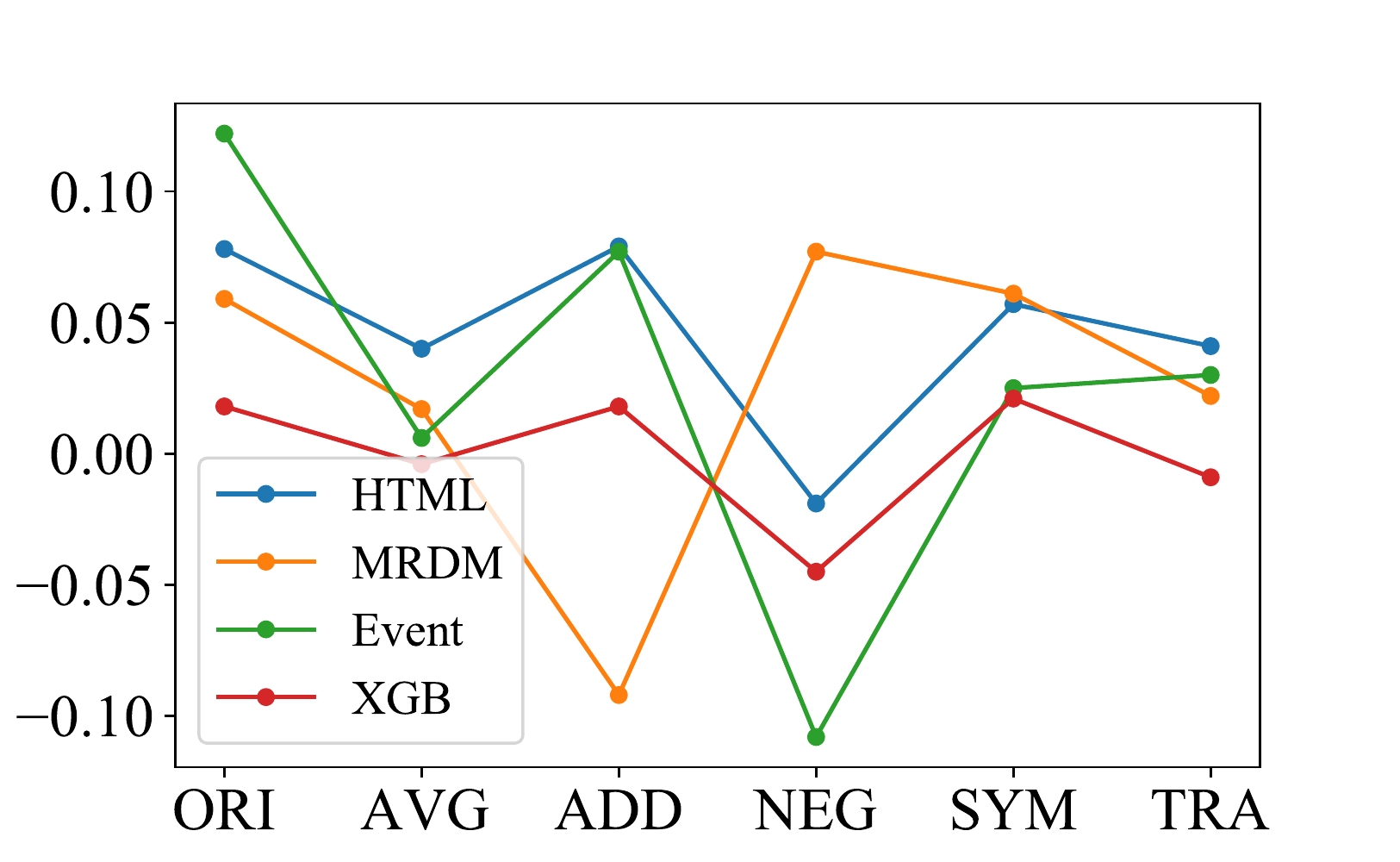}
    \caption{The ablation study of \method in stock movement prediction on the average MCC over periods.}
    \label{fig:ablation}
\end{figure} 

\noindent\textbf{Ablation Study.} We show the ablation results in stock movement prediction of four transformations in Figure \ref{fig:ablation}. We find that evaluations on the \method lead to significant performance decay for most settings compared to the original performance, which illustrates the individual influence of transformations. In particular, we show that models usually underperform when evaluating the \emph{negation transformation}, with the exception of MRDM. It suggests that current models lack the ability to provide non-contradictory predictions.

\section{Conclusion}

We proposed \method, an evaluation tool that assesses the trustworthiness of financial forecasting models in addition to their accuracy. Results on \method show that (1) the consistency of state-of-the-art models falls significantly short of expectations when applied to stock movement prediction; (2) predictions with such a low logical consistency can lead to severe consequences, as evidenced by poor performance in a trading simulation test. Our empirical results highlight the importance of perceiving such concerns when developing and evaluating text-based financial models, and we release our dataset for facilitating future research. Despite this, how to evaluate the consistency of large-scale language models (LLMs) is still an open question towards the financial forecasting task.


\section*{Limitation}

While our pipeline is designed to be applicable to any financial text dataset, the evaluation dataset is transformed solely on earnings conference calls. We will expand the scope of experiments to include other financial text sources such as news articles and social media posts. Finally, the current trading simulation does not take transaction costs into account. Going forward it will be necessary to consider more sophisticated trading policies.


\section*{Ethics Statement}
This paper honors the ACL Code of Ethics. The dataset used in the paper does not contain any private information. All annotators have received enough labor fees corresponding to their amount of annotated instances. The code and data are open-sourced under the CC-BY-NC-SA license.

\section*{Acknowledgements}
We would like to thank anonymous reviewers for their insightful comments and suggestions to help improve the paper. This publication has emanated from research conducted with the financial support of the Pioneer and ``Leading Goose" R\&D Program of Zhejiang under Grant Number 2022SDXHDX0003, the 72nd round of the Chinese Post-doctoral Science Foundation project 2022M722836, and the financial support from the rxhui.com company. Yue Zhang is the corresponding author.

\bibliography{custom}
\bibliographystyle{acl_natbib}

\clearpage
\appendix

\begin{table*}[]
\centering
\resizebox{\textwidth}{!}{%
\begin{tabular}{c|ccccc|ccccc|ccccc}
\hline
 &
  \multicolumn{5}{c|}{ACC} &
  \multicolumn{5}{c|}{F1} &
  \multicolumn{5}{c}{MCC} \\
\multirow{-2}{*}{} &
  3 &
  7 &
  15 &
  30 &
  Avg &
  3 &
  7 &
  15 &
  30 &
  Avg &
  3 &
  7 &
  15 &
  30 &
  Avg \\ \hline
\rowcolor[HTML]{EFEFEF} 
HTML &
  0.442 &
  0.531 &
  0.566 &
  0.646 &
  \textit{0.546} &
  0.571 &
  0.619 &
  0.713 &
  0.780 &
  \textit{0.671} &
  0.052 &
  0.056 &
  0.032 &
  0.175 &
  \textit{0.078} \\
HTML-Add &
  0.407 &
  0.540 &
  0.584 &
  0.619 &
  {\color[HTML]{000000} \textbf{0.538}} &
  0.579 &
  0.662 &
  0.715 &
  0.726 &
  0.671 &
  0.000 &
  0.085 &
  0.104 &
  0.127 &
  0.079 \\
HTML-Neg &
  0.602 &
  0.522 &
  0.416 &
  0.363 &
  {\color[HTML]{000000} \textbf{0.476}} &
  0.737 &
  0.625 &
  0.522 &
  0.532 &
  {\color[HTML]{000000} \textbf{0.604}} &
  0.089 &
  0.073 &
  -0.113 &
  -0.123 &
  {\color[HTML]{000000} \textbf{-0.019}} \\
HTML-Sym &
  0.425 &
  0.522 &
  0.566 &
  0.637 &
  {\color[HTML]{000000} \textbf{0.538}} &
  0.564 &
  0.620 &
  0.684 &
  0.776 &
  {\color[HTML]{000000} \textbf{0.661}} &
  0.004 &
  0.036 &
  0.066 &
  0.123 &
  {\color[HTML]{000000} \textbf{0.057}} \\
HTML-Tra &
  0.425 &
  0.522 &
  0.549 &
  0.637 &
  {\color[HTML]{000000} \textbf{0.533}} &
  0.552 &
  0.609 &
  0.671 &
  0.776 &
  {\color[HTML]{000000} \textbf{0.652}} &
  -0.016 &
  0.037 &
  0.021 &
  0.123 &
  {\color[HTML]{000000} \textbf{0.041}} \\
HTML-Avg &
  0.465 &
  0.527 &
  0.529 &
  0.564 &
  {\color[HTML]{000000} \textbf{0.521}} &
  0.608 &
  0.629 &
  0.648 &
  0.703 &
  {\color[HTML]{000000} \textbf{0.647}} &
  0.019 &
  0.058 &
  0.019 &
  0.063 &
  {\color[HTML]{000000} \textbf{0.040}} \\ \hline
\rowcolor[HTML]{EFEFEF} 
MRDM &
  0.504 &
  0.513 &
  0.584 &
  0.619 &
  \textit{0.555} &
  0.541 &
  0.663 &
  0.722 &
  0.754 &
  \textit{0.670} &
  0.079 &
  0.007 &
  0.107 &
  0.044 &
  \textit{0.059} \\
MRDM-Add &
  0.416 &
  0.496 &
  0.434 &
  0.496 &
  {\color[HTML]{000000} \textbf{0.460}} &
  0.507 &
  0.655 &
  0.289 &
  0.627 &
  {\color[HTML]{000000} \textbf{0.520}} &
  -0.079 &
  -0.073 &
  -0.073 &
  -0.145 &
  {\color[HTML]{000000} \textbf{-0.092}} \\
MRDM-Neg &
  0.619 &
  0.513 &
  0.434 &
  0.381 &
  {\color[HTML]{000000} \textbf{0.487}} &
  0.746 &
  0.667 &
  0.600 &
  0.539 &
  {\color[HTML]{000000} \textbf{0.638}} &
  0.153 &
  0.161 &
  -0.018 &
  0.013 &
  0.077 \\
MRDM-Sym &
  0.425 &
  0.531 &
  0.584 &
  0.628 &
  {\color[HTML]{000000} \textbf{0.542}} &
  0.504 &
  0.683 &
  0.697 &
  0.753 &
  {\color[HTML]{000000} \textbf{0.659}} &
  -0.067 &
  0.101 &
  0.111 &
  0.100 &
  0.061 \\
MRDM-Tra &
  0.398 &
  0.504 &
  0.575 &
  0.637 &
  {\color[HTML]{000000} \textbf{0.529}} &
  0.521 &
  0.663 &
  0.727 &
  0.776 &
  0.672 &
  -0.105 &
  -0.037 &
  0.108 &
  0.123 &
  {\color[HTML]{000000} \textbf{0.022}} \\
MRDM-Avg &
  0.465 &
  0.511 &
  0.507 &
  0.535 &
  {\color[HTML]{000000} \textbf{0.504}} &
  0.569 &
  0.667 &
  0.578 &
  0.674 &
  {\color[HTML]{000000} \textbf{0.622}} &
  -0.024 &
  0.038 &
  0.032 &
  0.023 &
  {\color[HTML]{000000} \textbf{0.017}} \\ \hline
\rowcolor[HTML]{EFEFEF} 
Event &
  0.416 &
  0.522 &
  0.593 &
  0.637 &
  \textit{0.542} &
  0.582 &
  0.682 &
  0.736 &
  0.776 &
  \textit{0.694} &
  0.078 &
  0.097 &
  0.189 &
  0.123 &
  \textit{0.122} \\
Event-Add &
  0.425 &
  0.522 &
  0.575 &
  0.637 &
  {\color[HTML]{000000} \textbf{0.540}} &
  0.558 &
  0.671 &
  0.652 &
  0.745 &
  {\color[HTML]{000000} \textbf{0.657}} &
  -0.007 &
  0.044 &
  0.116 &
  0.157 &
  {\color[HTML]{000000} \textbf{0.077}} \\
Event-Neg &
  0.531 &
  0.478 &
  0.381 &
  0.381 &
  {\color[HTML]{000000} \textbf{0.442}} &
  0.686 &
  0.638 &
  0.545 &
  0.539 &
  {\color[HTML]{000000} \textbf{0.602}} &
  -0.151 &
  -0.049 &
  -0.246 &
  0.013 &
  {\color[HTML]{000000} \textbf{-0.108}} \\
Event-Sym &
  0.416 &
  0.504 &
  0.593 &
  0.628 &
  {\color[HTML]{000000} \textbf{0.535}} &
  0.571 &
  0.671 &
  0.726 &
  0.767 &
  {\color[HTML]{000000} \textbf{0.684}} &
  0.003 &
  -0.092 &
  0.138 &
  0.051 &
  {\color[HTML]{000000} \textbf{0.025}} \\
Event-Tra &
  0.416 &
  0.513 &
  0.566 &
  0.628 &
  {\color[HTML]{000000} \textbf{0.531}} &
  0.577 &
  0.675 &
  0.707 &
  0.767 &
  {\color[HTML]{000000} \textbf{0.681}} &
  0.025 &
  0.004 &
  0.042 &
  0.051 &
  {\color[HTML]{000000} \textbf{0.030}} \\
Event-Avg &
  0.447 &
  0.504 &
  0.529 &
  0.569 &
  {\color[HTML]{000000} \textbf{0.512}} &
  0.598 &
  0.663 &
  0.658 &
  0.705 &
  {\color[HTML]{000000} \textbf{0.656}} &
  -0.032 &
  -0.023 &
  0.013 &
  0.068 &
  {\color[HTML]{000000} \textbf{0.006}} \\ \hline
\rowcolor[HTML]{EFEFEF} 
XGB &
  0.434 &
  0.487 &
  0.584 &
  0.558 &
  \textit{0.515} &
  0.448 &
  0.500 &
  0.641 &
  0.653 &
  \textit{0.561} &
  -0.093 &
  -0.027 &
  0.147 &
  0.043 &
  \textit{0.018} \\
XGB-Add &
  0.398 &
  0.504 &
  0.593 &
  0.575 &
  0.518 &
  0.433 &
  0.533 &
  0.657 &
  0.676 &
  0.575 &
  -0.156 &
  0.006 &
  0.160 &
  0.064 &
  0.018 \\
XGB-Neg &
  0.549 &
  0.469 &
  0.398 &
  0.451 &
  {\color[HTML]{000000} \textbf{0.467}} &
  0.622 &
  0.444 &
  0.404 &
  0.492 &
  {\color[HTML]{000000} \textbf{0.490}} &
  0.062 &
  -0.064 &
  -0.187 &
  0.011 &
  {\color[HTML]{000000} \textbf{-0.045}} \\
XGB-Sym &
  0.434 &
  0.496 &
  0.575 &
  0.566 &
  0.518 &
  0.448 &
  0.513 &
  0.636 &
  0.662 &
  0.565 &
  -0.093 &
  -0.010 &
  0.127 &
  0.058 &
  0.021 \\
XGB-Tra &
  0.469 &
  0.540 &
  0.558 &
  0.531 &
  0.524 &
  0.318 &
  0.581 &
  0.638 &
  0.658 &
  {\color[HTML]{000000} \textbf{0.549}} &
  -0.115 &
  0.076 &
  0.078 &
  -0.075 &
  {\color[HTML]{000000} \textbf{-0.009}} \\
XGB-Avg &
  0.462 &
  0.502 &
  0.531 &
  0.531 &
  {\color[HTML]{000000} \textbf{0.507}} &
  0.456 &
  0.518 &
  0.584 &
  0.622 &
  {\color[HTML]{000000} \textbf{0.545}} &
  -0.076 &
  0.002 &
  0.045 &
  0.014 &
  {\color[HTML]{000000} \textbf{-0.004}} \\ \hline
\end{tabular}%
}
\caption{Ablation study results of four different types of logical transformation based on the fine-tuned forecasting models. Compared to the original results, the decreased performance is presented in \textbf{bold}.}
\label{tab:my-table5}
\end{table*}

\section{Transitive Consistency}
\label{sec:appendix}

\textbf{Example.} We show an example to understand better the motivation for using Transitive Consistency when measuring the consistency of FinNLP models. Given \textit{"Nektar Therapeutics gave investors strong confidence after Earnings Conference Call on March 1, 2017, and its stock price soared 79.43\% in the following month."}. As a leading company in the same Sector (Health Care), Johnson \& Johnson (JNJ) was also affected by this and increased by 1.91\% over the same period, which confirmed the rationality of selecting transitive consistency as one of the measurement methods.

\section{Full Ablation Results}
\label{sec:appendix}

\begin{table*}[]
\centering
\resizebox{\textwidth}{!}{%
\begin{tabular}{c|cc|c|cc|c|cc}
\hline
Strategy       & Profit Ratio & Sharpe Ratio & Transformations & Profit Ratio & Sharpe Ratio & Transformations & Profit Ratio & Sharpe Ratio  \\ \hline
HTML-Original           & 3.752        & 0.266        & 
HTML-ADD            & 2.282$\downarrow$       & 0.125$\downarrow$       &
HTML-NEG            & 3.720$\downarrow$       & 0.263$\downarrow$       \\
HTML+FinTrust            & 3.359$\downarrow$       & 0.229$\downarrow$       &
HTML-SYM            & 3.720$\downarrow$       & 0.263$\downarrow$       &
HTML-TRA            & 3.713$\downarrow$       & 0.263$\downarrow$       \\ \hline

Event-Original          & 3.720        & 0.263        &
Event-ADD            & 3.646$\downarrow$       & 0.256$\downarrow$       &
Event-NEG            & 3.347$\downarrow$       & 0.226$\downarrow$       \\
Event+FinTrust            & 3.535$\downarrow$       & 0.245$\downarrow$       &
Event-SYM            & 3.494$\downarrow$       & 0.241$\downarrow$       &
Event-TRA            & 3.652$\downarrow$       & 0.256$\downarrow$       \\ \hline

MRDM-Original           & 3.495        & 0.241        &
MRDM-ADD            & 0.605$\downarrow$       & -0.026$\downarrow$       &
MRDM-NEG            & 3.512$\uparrow$       & 0.243$\uparrow$       \\
MRDM+FinTrust            & 2.384$\downarrow$       & 0.138$\downarrow$       &
MRDM-SYM            & 1.674$\downarrow$       & 0.070$\downarrow$       &
MRDM-TRA            & 3.743$\uparrow$       & 0.266$\uparrow$      \\ \hline

XGB-Original            & -0.515       & -0.126       &
XGB-ADD            & 0.972$\uparrow$       & 0.006$\uparrow$       &
XGB-NEG            & -0.833$\downarrow$      & -0.067$\uparrow$      \\
XGB+FinTrust            & 0.296$\uparrow$       & -0.032$\uparrow$       & 
XGB-SYM            & -0.072$\uparrow$      & -0.087$\uparrow$      &
XGB-TRA            & 1.118$\uparrow$      & 0.020$\uparrow$       \\ \hline
\end{tabular}%
}
\caption{The ablation study of the trading simulation based on text-based fine-tuned
forecasting models.}
\label{tab:my-table6}
\end{table*}

We report the ablation study results of four different types of logical transformation based on the fine-tuned forecasting models in Table \ref{tab:my-table5}. We use \textit{italics} to indicate the performance before consistency transformation, use \textbf{bold} to express the performance that has been reduced after consistency transformation, and do not deal with other parts that have not decreased, for the convenience of readers. All detailed return changes in trading simulation based on text-based fine-tuned forecasting models are also shown in Table \ref{tab:my-table6}. "+\method" means the average impact of the four transformations.

\section{Additional experimental details}
\label{sec:appendix}

The model settings involved in the paper are all aligned with the parameters and training details described in the corresponding article \citet{yang2020html,Qin19,Ding15,chen2015xgboost}. The total computational budget is about 50 GPU hours, using a GeForce RTX 3090. All models use the highest performance among ten repeated experiments using different seeds and ensure reproducibility.

\end{document}